\documentclass[10pt,twocolumn,letterpaper]{article}

\usepackage{iccv}
\usepackage{times}
\usepackage{epsfig}
\usepackage{graphicx}
\usepackage{amsmath}
\usepackage{amssymb}
\usepackage{mathtools}

\newcommand{\ignore}[1]{}

\newcommand{\ba}{\begin{array}}
\newcommand{\ea}{\end{array}}
\newcommand{\bc}{\begin{center}}
\newcommand{\ec}{\end{center}}
\newcommand{\be}{\begin{enumerate}}
\newcommand{\ee}{\end{enumerate}}
\newcommand{\bea}{\begin{eqnarray}}
\newcommand{\eea}{\end{eqnarray}}
\newcommand{\beas}{\begin{eqnarray*}}
\newcommand{\eeas}{\end{eqnarray*}}
\newcommand{\beq}{\begin{equation}}
\newcommand{\eeq}{\end{equation}}
\newcommand{\bfig}{\begin{figure}}
\newcommand{\efig}{\end{figure}}
\newcommand{\bi}{\begin{itemize}}
\newcommand{\ei}{\end{itemize}}
\newcommand{\bpic}{\begin{picture}}
\newcommand{\epic}{\end{picture}}
\newcommand{\btabular}{\begin{tabular}}
\newcommand{\etabular}{\end{tabular}}
\newcommand{\btable}{\begin{table}}
\newcommand{\etable}{\end{table}}

\newcommand{\es}{\vfill
                 \rule[-6mm]{170mm}{0.7mm} \\
                 \redw{{\tiny 
		  \hfill S-\theslide}}
                 \end{slide}}

\newcommand{\matxx}[1]{{\mathtt #1}}
\newcommand{\vecXX}[1]{{\mathbf {#1}}}
\newcommand{\vecYY}[1]{{\boldsymbol {#1}}}

\newcommand{\argmin}{\operatornamewithlimits{arg\ min}}

\def \crossprod {\times}

\providecommand{\etal}{{\em et~al.}}

\def \hbar {{\bar{h}}}

\def \vecb {{\vecXX{b}}}

\def \vece {{\vecXX{e}}}

\def \vecp {{\vecXX{p}}}

\def \vecr {{\vecXX{r}}}

\def \vecx {{\vecXX{x}}}

\def \vecomega {{\vecYY{\omega}}}

\def \matG {{\matxx{G}}}
\def \matH {{\matxx{H}}}
\def \matI {{\matxx{I}}}
\def \matJ {{\matxx{J}}}
\def \matK {{\matxx{K}}}

\def \matR {{\matxx{R}}}

\def \matV {{\matxx{V}}}
\def \matW {{\matxx{W}}}

\def \vecxdot {\dot{\vecXX{x}}}

\newcommand{\diff}[2]{ \frac{d#1}{d#2} }

\newcommand{\pdiff}[2]{ \frac{\partial#1}{\partial#2} }

\def \inv {^{-1}}

\newcommand{\bracks}[1]{ \left({#1}\right)}

\makeatletter
\renewcommand*\env@matrix[1][*\c@MaxMatrixCols c]{%
  \hskip -\arraycolsep
  \let\@ifnextchar\new@ifnextchar
  \array{#1}}
\makeatother

\newcommand{\RR}{\mathbb{R}}

\newcommand{\so}[1]{\ensuremath{{\mathfrak{so}(#1)}}}

\newcommand{\norm}[1]{\lVert#1\rVert}

\graphicspath{{./img/}}

\usepackage[pagebackref=true,breaklinks=true,letterpaper=true,colorlinks,bookmarks=false]{hyperref}

\iccvfinalcopy 

\def\iccvPaperID{4} 

\ificcvfinal\fi
\begin{document}

\title{Semantic Texture for Robust Dense Tracking}

\author{Jan Czarnowski\\
{\tt\small j.czarnowski15@imperial.ac.uk}
\and
Stefan Leutenegger\\
{\tt\small s.leutenegger@imperial.ac.uk}
\and
Andrew J. Davison\\
{\tt\small a.davison@imperial.ac.uk}
\and
Dyson Robotics Laboratory at Imperial College, Dept. of Computing, Imperial
College London, UK
}

\maketitle

\begin{abstract}
We argue that robust dense SLAM systems can make valuable use of the
layers of features coming from a standard CNN as a pyramid of `semantic texture' which is suitable
for dense alignment while being much more robust to nuisance factors
such as lighting than raw RGB values.
We use a straightforward Lucas-Kanade formulation of image alignment,
with a schedule of iterations  over the
coarse-to-fine levels of  a pyramid, and simply 
replace the usual image pyramid
by the hierarchy of convolutional feature maps from a pre-trained CNN.
The resulting dense alignment performance is much more
  robust to lighting and other variations, as we show by
  camera rotation tracking experiments on time-lapse
  sequences captured over many hours.
  Looking towards the future of scene representation for real-time visual SLAM, we further demonstrate that a selection using simple criteria of a small number of the total set of features output by a CNN gives just as accurate but much more efficient tracking performance.
  
\end{abstract}

\section{Introduction} 
Dense visual SLAM~\cite{Newcombe:etal:ICCV2011} involves incrementally
reconstructing the whole appearance of a scene rather reducing it to sparse
features, and this approach is fundamental if we want scene models
with general, generative representations.
Tracking is achieved via whole image alignment of live images with the
reprojected dense texture of the reconstruction. However, using raw
RGB values in persistent dense scene representations over long time
periods is problematic because of
their strong dependence on lighting and other imaging factors, causing
image to model alignment to fail.
While
one thread of research to mitigate this involves getting closer to the 
real physics of the world by modelling lighting and surface
reflectance in detail, this is very computationally challenging in
real-time. The alternative, which we pursue here, is to give up on
representing light intensity directly in scene representations, and
instead to use transformations which capture the information important
for tracking but are invariant to nuisance factors such as lighting.

A long term goal in SLAM is to
replace the raw geometry and appearance information in a 3D scene map
by high level semantic entities such as walls, furniture, and
objects. This is an approach being followed by many groups who are
working on semantic labelling and object recognition within SLAM~\cite{Hermans:etal:ICRA2014,Valentin:etal:CVPR2014,McCormac:etal:ICRA2017},
driving towards systems capable of scene mapping at the level of
nameable entities (a direction pointed to by the SLAM++ system~\cite{Salas-Moreno:etal:CVPR2013}).

What we argue here, and make the first experimental steps to
demonstrate, is that there is a range of very useful levels of
representation for mapping and tracking 
in between raw pixel values and object level semantics. The explosion
of success in computer vision by Convolutional Neural Networks, and
work on investigating and visualising the levels of features they
generate in a variety of vision tasks
(e.g. \cite{Simonyan:etal:ARXIV2013}), has revealed a straightforward
way to get at these representations as the outputs of successive
levels of convolutional feature banks in a CNN.

In this paper we demonstrate that dense alignment, the most
fundamental component of dense SLAM, can be formulated simply to make
use not of a standard image pyramid, but the responses of the layers
of a standard CNN trained for classification; and that this leads to
much more robust tracking in the presence of difficult conditions such
as extreme lighting changes. We demonstrate our results in a pure
rotation SLAM system, where long term tracking against
keyframes is achieved over the lighting variations during a whole day. We perform detailed experiments on the convergence basins of
alignment at all pyramid levels, comparing CNN
pyramids against raw RGB and dense SIFT. We also show that we achieve just
as good performance with small percentages of CNN features chosen
using simple criteria of persistence and texturedness, pointing to
highly efficient real-time solutions in the near future.

\section{Background}

The
canonical approach to dense image alignment is due to Lucas and Kanade \cite{Lucas:Kanade:IJCAI1985},
and is laid out in more detail in the review papers by Baker and
Matthews~\cite{Baker:Matthews:IJCV2004}.
In the LK algorithm, given an initial hypothesis
for the transformation, the current `error' between the images is calculated by
adding the squared difference in greyscale or RGB values between all pairs of
pixels from the two images which are brought into correspondence by this
transformation. The derivative of this error with respect to the transformation
parameters is determined, and a Newton step is taken to reduce the error.
It has
been widely demonstrated that the correct transformation is found at the minimum
of the error function after a number of iterations.

The performance of dense alignment can be measured along several axes. Two are
the accuracy of the final alignment and the speed of convergence, but generally
more important are the size of the basin of convergence and the robustness to unmodelled effects such as lighing changes.  Both the speed and
basin of convergence of LK alignment are improved by the use of image pyramids.
Before alignment, both images are successively downsampled to produce a stack of
images with decreasing resolution. Alignment then proceeds by performing a
number of iterations at each level, starting with the lowest resolution versions
which retain only low frequency detail but allow fast computation, and ending back at the original versions where only a few of the
most expensive iterations are needed.

Here we replace this
downsampling pyramid in LK by the output of the convolutional layers of an
off-the-shelf VGG Convolutional Neural Network trained for classification.
The early layers of a CNN
are well known to be generic, regardless of the final task it was
trained for, as shown in~\cite{Agarwal:etal:ICCV2015}. The later layers respond
to features which are increasingly complex and semantically meaningful, with more
invariance to transformations including illumination.  Now that the convolutional
layers of a CNN can comfortably be run in real-time on video input on desktop or
mobile GPUs, the simplicity of using them to build pyramids
for alignment and scene representation is very appealing.

Even though CNNs turn raw images into feature maps, we argue that aligning their
hierarchical responses is still a `dense' method, which is much more akin to
standard Lucas-Kanade alignment than image matching using sparse features.  The
convolutions produce a response at every image location. As we move
towards powerful real-time scene understanding systems which can deal with
complex scenes whose appearance and shape changes over short and long
timescales, we will need dense and fully generative representations, and the
ability to fuse and test live data against these at every time-step.
We believe
that there are many interesting levels of representation to find between raw
pixel values and human annotated `object-level' models, and that these
representations should be learned and optimised depending on what task needs to
be achieved, such as detecting whether something has changed in a scene.

Other non-CNN transformations of RGB have been attempted to
improve the performance of correspondence algorithms. There are many advantages
to generating a scale-space pyramid using non-linear diffusion
\cite{Perona:Malik:PAMI1990}, where edges are preserved while local noise is
smoothed away, although it is not clear how much this would help Lucas-Kanade
alignment. In stereo matching, there has been work such as that by
Hirschm{\"u}ller \etal~\cite{Hirschmuller:CVPR2007} which  compared three
representations: Laplacian of Gaussian (LoG), rank filter and mean filter. All
of these approaches help with viewpoint and lighting changes.  In
face fitting using Lucas-Kanade alignment, Antonakos~\etal
\cite{Antonakos:etal:IP2015} evaluated nine different dense hand-designed
feature transformations and found SIFT and HOG to be the most powerful.

There is a growing body of literature that uses convolutional networks to compare image patches, register camera pose and compute optical flow between images~\cite{Fischer:etal:arXiv2015,Zagoruyko:Komodakis:CORR2015,Luo:etal:CVPR2016,Kendall:etal:CORR2015}. Such methods use networks trained end-to-end, and can output robust estimates in challenging conditions but fail to deliver the accuracy of model-based approaches. The training set is trusted to encompass all possible variations, a hard condition to meet in the real world. Instead of performing iterative refinement those methods produce a one-shot estimate. In FlowNet, in particular, the optical flow result must still be optimised with a standard variational scheme (TV-L1). Our approach bridges the gap between these two paradigms, delivering both the accuracy of online optimisation and the robustness of learned features. 

\section{Dense Image Alignment}
\subsection{Direct Per-Pixel Cost Function}
\label{sec:direct_cost_fun}
Image alignment (registration) requires of moving and deforming a
constant template image to find the  best match with a reference image.  At the
heart of image alignment lies a generative warp model, which is parameterised
to represent the degrees of freedom of relative camera--scene motion. Alignment
proceeds by iteratively finding better sets of parameters to warp the images
into correspondence.

To assess correspondence we must define a measure of image
similarity. In the standard form of LK, sum of squared differences (SSD) is
used. This gives the following objective function:
\begin{equation}
  \label{eq:cost_fun}
  \argmin_\vecp \sum_\vecx || \matI_r(\matW(\vecx;\vecp)) -
  \matI_t(\vecx) ||^2
  ~,
\end{equation}
where $\matI_r$ is the reference image, $\matI_t$ is the template image, $x$ is
an image pixel location and $\matW(\vecx;\vecp): \RR^2 \rightarrow \RR^2$ is the
generative warp model. The vector of warp parameters we aim to solve for is
$\vecp$.

One way to find the optimal parameters is simply `exhaustive search',
evaluating the cost function over a full range of quantised parameter
combinations and selecting the minimum. Visualising the cost surface produced is insightful; Figure~\ref{fig:cost_landscape} presents a 2
degree-of-freedom example for images related by pan-tilt camera rotation. 
\bfig
  \centering
  \includegraphics[width=\columnwidth,clip,trim=4cm 4cm 4cm
  4cm]{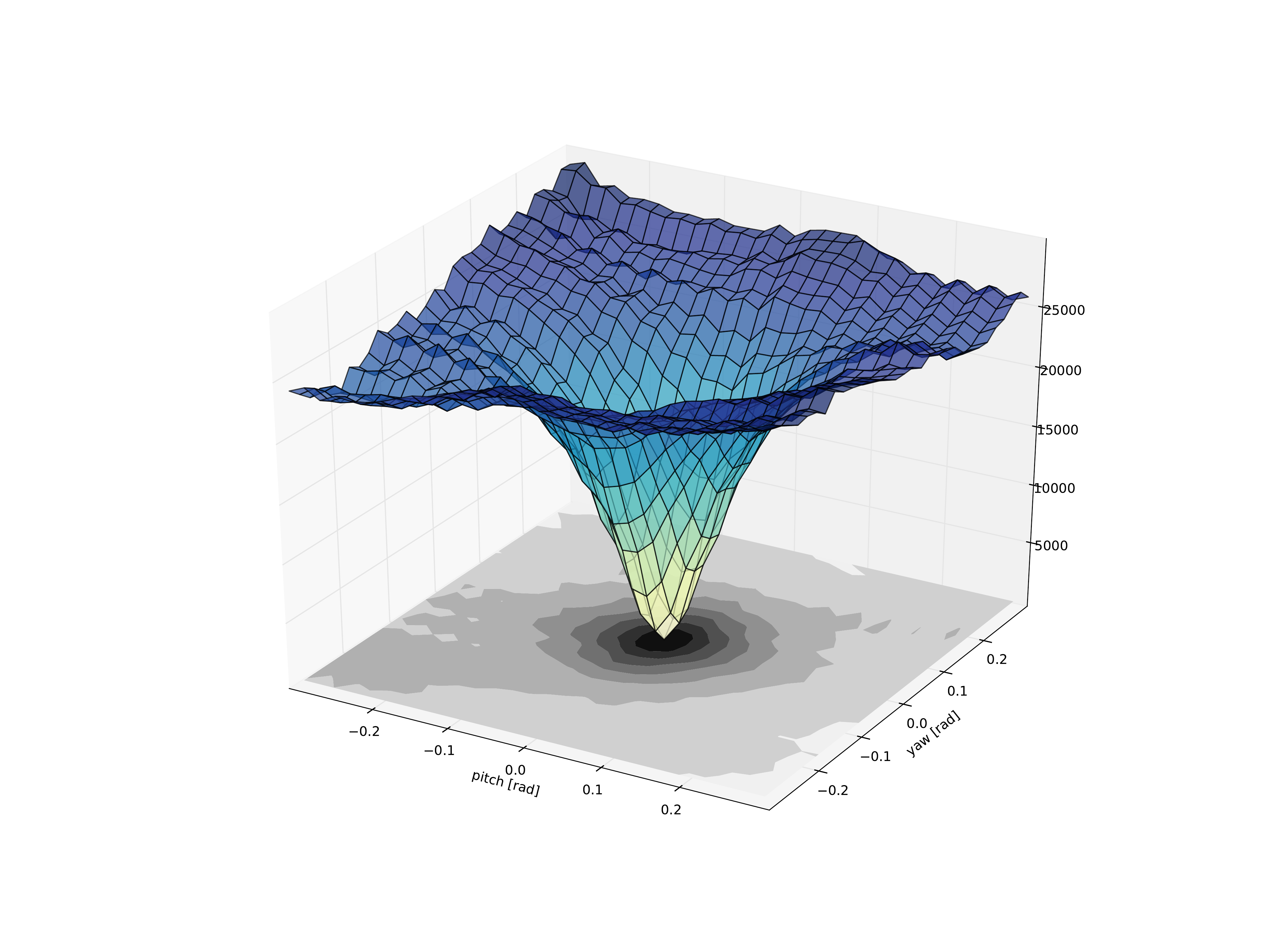}
  \caption{Example 2DoF image alignment cost function for
    pan-tilt camera rotation, with a clear minimum and
    smooth bowl.}
  \label{fig:cost_landscape}
\vspace{2mm}\hrule 
\efig
The clear bowl shape of this surface shows that we can do something
much more efficient than exhaustive search as long as we start from a
set of parameters close enough to correct alignment by following the
gradient to the minumum.

\subsection{Lucas-Kanade Algorithm}
\label{sec:lucas_kanade}
In LK alignment~\cite{Lucas:Kanade:IJCAI1985}, at each iteration we use the
gradient of the cost function in Equation~\ref{eq:cost_fun} with respect to the
warp parameter vector $\vecp$, and determine a parameter update $\Delta \vecp$
which takes us closer to the minimum of the function. 

A more efficient version of this algorithm, which allows for pre-computation of
the system Jacobian and Hessian was proposed in~\cite{Baker:Matthews:IJCV2004}.
The trick consists of swapping the roles of the reference and template image
and optimising for an update warp composed with the current warp estimate. The
modified cost function has the form:
\beq
  \label{eq:ik_lk}
  \Delta \vecp = \argmin_{\Delta \vecp} \sum_x \norm{\matI_t(\matW(\vecx;\Delta \vecp)) -
  \matI_r(\matW(\vecx;\vecp))}^2
  ~,
\eeq
with the following update rule:
\beq
  \matW(\vecx;\vecp) \leftarrow \matW(\vecx;\vecp) \circ \matW(\vecx;\Delta \vecp)^{-1}
\eeq
Linearizing Equation~\ref{eq:ik_lk} leads to a  closed form solution:

\beq
  \Delta \vecp = \bracks{\sum_\vecx \matJ^T\matJ}\inv \sum_\vecx \matJ^T\vecr
  ~,
\eeq
where:
\beq
  \matJ = -\nabla\matI_t \diff{\matW(\vecx,\vecp)}{\vecp} \bigg|_{\vecp=0}
\eeq

\beq
  \vecr = \matI_r(\matW(\vecx,\vecp)) - \matI_t(\vecx)
\eeq 
The shape of the convergence basin heavily depends on the image content: amount
and type of texture and ambiguities present in the image. Since the cost
landscape is usually locally convex in vincinity of the real translation, a
good initialization is required for the optimization to successfuly converge to
the correct solution.

\subsection{Coarse-to-Fine Alignment}
To increase the size of the convergence basin, the original image can be smoothed
using Gaussian blur, which simplifies the cost surface through removing the
details in the image~\cite{Lucas:Kanade:IJCAI1981}. The Gaussian smoothed image
is usually downsampled since it does not cause loss of information and reduces
the number of pixels to be processed. Since a pixel in the downsampled image
corresponds to multpile pixels in the original image, the distance between the
two is shortened, which increases the size of the basin of convergence. For the
same reason, the accuracy of the alignment performed on smaller version of the
images is also reduced.

The common approach is to start the optimization using highly downsampled and
blurred images to obtain an initial estimate and refine it using progressively
more detailed versions of the images. This can be perceived as a pyramid of
degraded versions of the image.

\section{Replacing the Pyramid with CNN Feature Maps} 
\begin{figure}
  \centering
  \def\svgwidth{\columnwidth}
  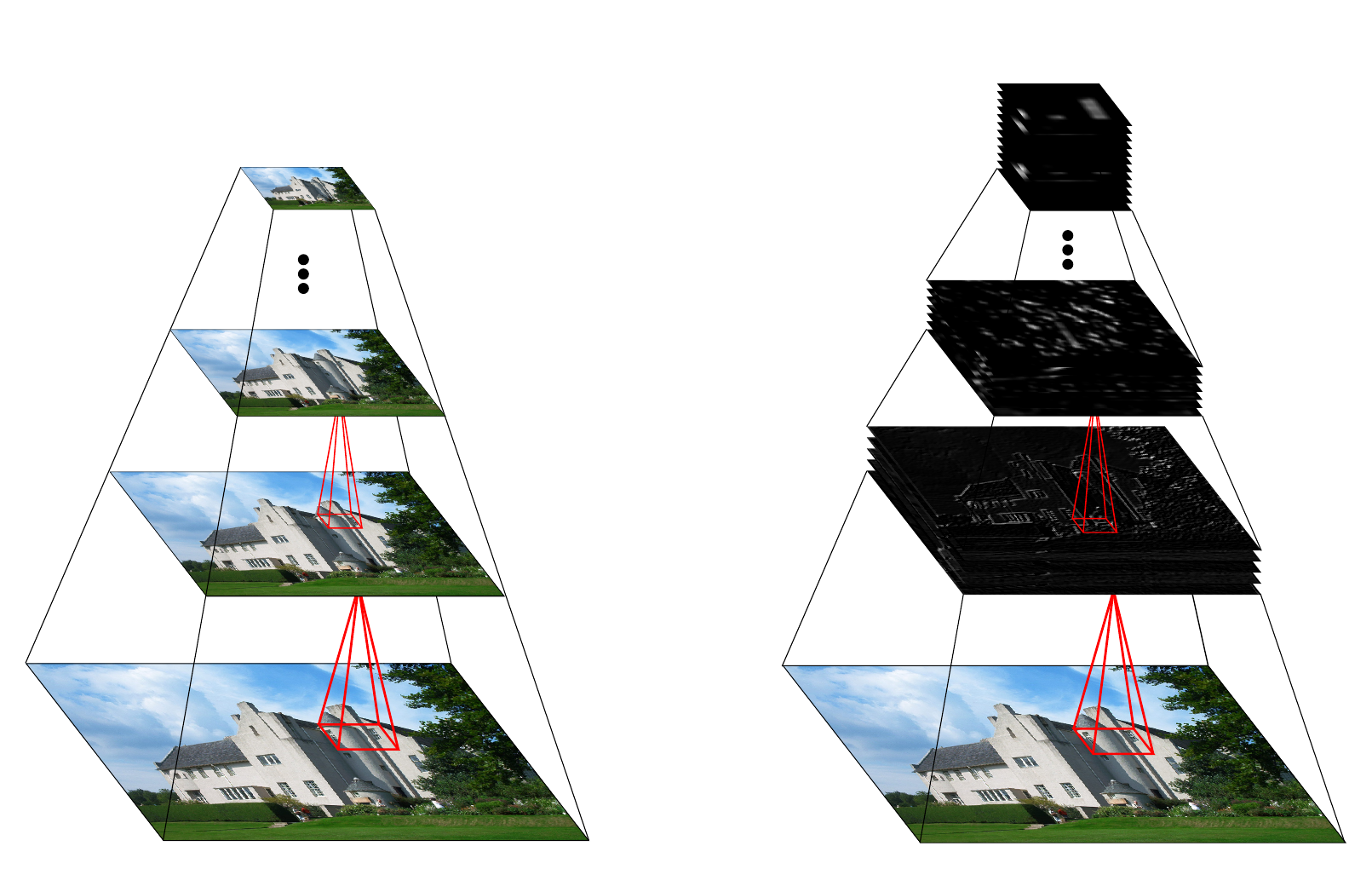
  \caption{A comparison between a Gaussian pyramid (left) and proposed
    conceptual pyramid (right).
  }
  \label{fig:pyr}
  \vspace{2mm}\hrule 
\end{figure}

Outputs of the convolutions in
standard classification CNN's form a pyramid similar to the ones used in RGB
based LK image alignment (Figure~\ref{fig:pyr}). Consecutive layers of
such pyramid encode increasingly more semantic information, starting from
simple geometrical filters in the first layers, leading to more complex
concepts. 

Each level of the pyramid takes the form of a multi-channel image
(tensor), where each channel contains responses to a certain learned
convolutional filter. We propose to align the volumes in a coarse-to-fine
manner, starting from the highest, low resolution layers and progressing down
to the lower, geometrical layers to refine alignment.

Section~\ref{sec:feature_extraction} presents the process of extracting the
feature pyramid out of an input RGB image. Following this, Section
(\ref{sec:volume_alignment}) describes how standard Lucas-Kanade alignment can
be applied to the proposed representation.

\subsection{Feature Extraction}
\label{sec:feature_extraction}

\bfig[h]
  \label{fig:net}
  \def\svgwidth{\columnwidth}
\begingroup%
  \makeatletter%
  \providecommand\color[2][]{%
    \errmessage{(Inkscape) Color is used for the text in Inkscape, but the package 'color.sty' is not loaded}%
    \renewcommand\color[2][]{}%
  }%
  \providecommand\transparent[1]{%
    \errmessage{(Inkscape) Transparency is used (non-zero) for the text in Inkscape, but the package 'transparent.sty' is not loaded}%
    \renewcommand\transparent[1]{}%
  }%
  \providecommand\rotatebox[2]{#2}%
  \ifx\svgwidth\undefined%
    \setlength{\unitlength}{393.97893951bp}%
    \ifx\svgscale\undefined%
      \relax%
    \else%
      \setlength{\unitlength}{\unitlength * \real{\svgscale}}%
    \fi%
  \else%
    \setlength{\unitlength}{\svgwidth}%
  \fi%
  \global\let\svgwidth\undefined%
  \global\let\svgscale\undefined%
  \makeatother%
  \begin{picture}(1,0.32930923)%
    \put(0,0){\includegraphics[width=\unitlength,page=1]{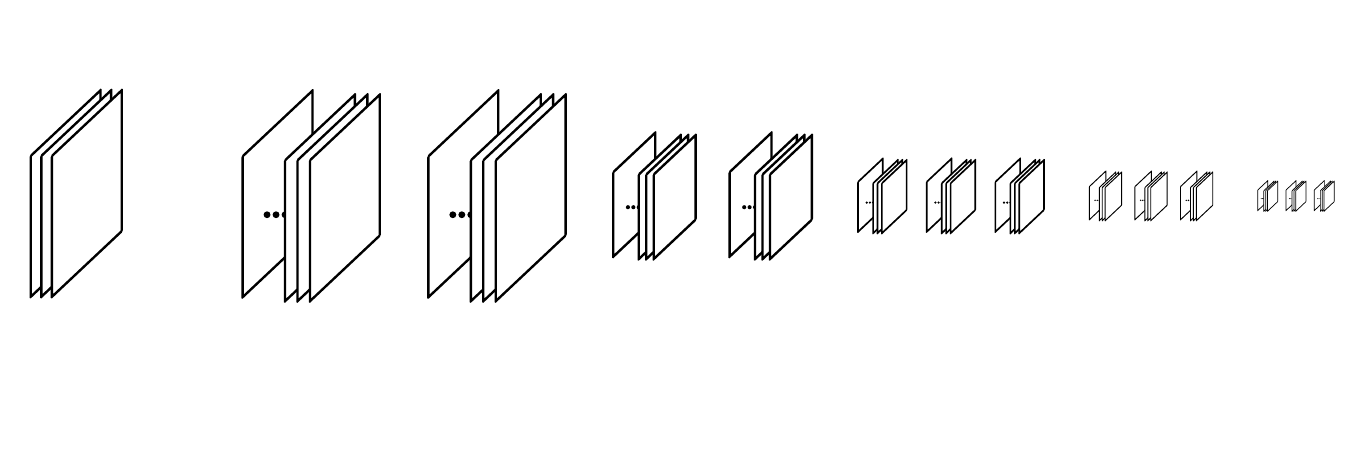}}%
    \put(0.02247989,0.08199732){\color[rgb]{0,0,0}\makebox(0,0)[lb]{\smash{3\\ }}}%
    \put(0.11244276,0.12032758){\color[rgb]{0,0,0}\makebox(0,0)[lb]{\smash{}}}%
    \put(0.20858425,0.06982406){\color[rgb]{0,0,0}\makebox(0,0)[lb]{\smash{64\\ }}}%
    \put(0.32972317,0.07219936){\color[rgb]{0,0,0}\makebox(0,0)[lb]{\smash{64\\ }}}%
    \put(0.44254863,0.09645029){\color[rgb]{0,0,0}\makebox(0,0)[lb]{\smash{128\\ }}}%
    \put(0.5339967,0.09645029){\color[rgb]{0,0,0}\makebox(0,0)[lb]{\smash{128\\ }}}%
    \put(0.66337373,0.11713833){\color[rgb]{0,0,0}\makebox(0,0)[lb]{\smash{256\\ }}}%
    \put(0.80933723,0.13326704){\color[rgb]{0,0,0}\makebox(0,0)[lb]{\smash{512\\ \\ }}}%
    \put(0.92522743,0.13395645){\color[rgb]{0,0,0}\makebox(0,0)[lb]{\smash{512\\ }}}%
    \put(0,0){\includegraphics[width=\unitlength,page=2]{network.pdf}}%
  \end{picture}%
\endgroup%

  \caption{A pyramid is created using successive learned convolutions, just as
  in a standard CNN classification network. We have used VGG-16 for our
  experiments}
\vspace{2mm}\hrule 
\efig

The pyramid is created through applying successive convolutions followed by a
nonlinear activation function, with occasional downsampling, very much like in
standard CNN's. The weights for convolutions are trained for an image
classification task. Although we believe that any classification CNN could be
used in our method, we have only tested weights from VGG-16 classification
network from~\cite{Simonyan:Zisserman:ICLR2015}. The network consists of 13
layers of convolutions. Figure~\ref{fig:net} presents the architecture of this
network.

\subsection{Volume Alignment}
\label{sec:volume_alignment}
For aligning volumes we use the inverse compositional cost function formulation
described in Section~\ref{sec:lucas_kanade}:

\beq
  \sum_\vecx \norm{\vece_\vecx(\Delta \vecp;\vecp)}^2 = \sum_\vecx
  \norm{\matV_t(\matW(\vecx;\Delta \vecp))-\matV_r(\matW(\vecx;\vecp))}^2
\eeq

Similar to normal images, we define a volume of depth $N$ as a function:
$\matV: \RR^2 \rightarrow \RR^N$. In order to solve for the parameter update
$\Delta p$, one first need to calculate the Jacobian and Hessian of this
nonlinear Least Squares System:

\beq
  \matJ_\vecx = \pdiff{\vece_\vecx}{\Delta \vecp} = 
  \begin{bmatrix} \pdiff{e_{\vecx,1}}{\Delta \vecp} \\ \vdots \\
  \pdiff{e_{\vecx,N}}{\Delta \vecp} \end{bmatrix} =
  \begin{bmatrix} \matJ_{\vecx,1} \\ \vdots \\ \matJ_{\vecx,N} \end{bmatrix}
\eeq

\beq
  \label{eq:Hsplit}
  \begin{split}
    \matH := \sum_\vecx \matJ_\vecx^T\matJ_\vecx &= \sum_\vecx \left[
    \matJ_{\vecx,1}^T \dots \matJ_{\vecx,N}^T \right]
  \begin{bmatrix} \matJ_{\vecx,1} \\ \vdots \\ \matJ_{\vecx,N} \end{bmatrix} =  \\
    &= \sum_\vecx \sum_{c = 1}^N \matJ_{\vecx,c}^T\matJ_{\vecx,c}
  \end{split}
\eeq

\beq
  \label{eq:jtrsplit}
  \begin{split}
    \vecb := 
    \sum_\vecx \matJ_\vecx^T \vece_\vecx &= \sum_\vecx \left[
    \matJ_{\vecx,1}^T \dots \matJ_{\vecx,N}^T \right] 
  \begin{bmatrix} \vece_{\vecx,1} \\ \vdots \\ \vece_{\vecx,N}^T \end{bmatrix} = \\
    &= \sum_\vecx \sum_{c = 1}^N \matJ_{\vecx,c}^T \vece_{\vecx, c}
  \end{split}
\eeq
The update is calculated using the normal equations:
\beq
\Delta \vecp = \matH\inv \vecb
\eeq

In order to perform alignment in real-time over big volumes we have 
implemented the optimization on GPU using NVIDIA CUDA. Each computational thread
calculates per-pixel values, which are later reduced to single matrices using
Equations~\ref{eq:Hsplit} and~\ref{eq:jtrsplit}.

\section{Live Spherical Mosaicing}
\label{sec:sphereslam}

\begin{figure}
  \centering
  \includegraphics[width=\columnwidth]{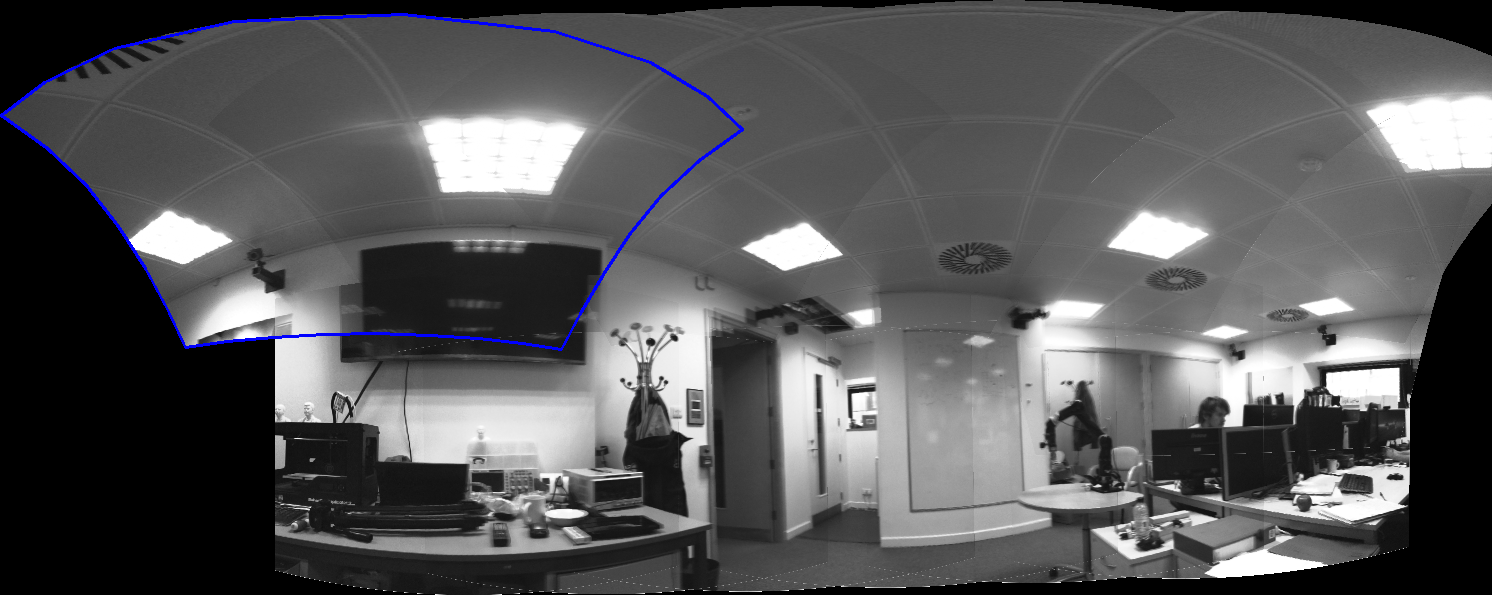}
  \caption{An example spherical panorama produced by tracking the camera
  rotation using CNN features}
  \label{fig:mosaic}
\vspace{2mm}\hrule 
\end{figure}

In order to test tracking using the proposed representation,
we have developed a real-time spherical mosaicing system similar to~\cite{Lovegrove:Davison:ECCV2010}. It tracks purely
rotational camera movement and produces a keyframe map, which is rendered into
a spherical panorama.
For purely rotational camera motion, the homography between two camera frames
with intrinsic matrix $\matK$, separated by rotation described by matrix
$\matR$ is given by: $\matK\matR\matK\inv$~\cite{Hartley:Zisserman:Book2004}.
Therefore, the generative warp function from Section~\ref{sec:direct_cost_fun}
has the form:
\beq
  \matW(\vecx;\vecomega) = \pi (\matK \matR(\vecomega) \matK^{-1}
  \vecxdot)
  ~,
\eeq
where $\vecxdot$ is the homogenized pixel location $\vecx$, and $\pi$ is the
projection function. We parametrize the incremental rotation $\matR(\vecomega)$
with $\vecomega \in R^3$. The incremental rotation described by vector
$\vecomega$ is mapped to the $\so3$ group elements via~\cite{Eade:Lie2014}:
\beq
\matR(\vecomega) = \exp (\vecomega^\crossprod)
~.
\eeq
The final cost function to be optimized takes the form:
\beq
  \sum_{x} ||\matV_t(\pi (\matK \matR(\vecomega) \matK^{-1} \vecxdot) -
  \matV_r(\matK\matR\matK^{-1})||^2
  ~.
\eeq
The keyframe map is projected onto a final compositioning surface. We use
spherical projection, which projects each keyframe onto a unit sphere, which is
later unrolled into a 2D surface for viewing. An example mosaic produced by our
system is presented in Figure~\ref{fig:mosaic}. We use an image of size
$224\times224$ to extract features and align all of the levels.

The system runs in real-time 15-20 frames per second using the whole pyramid.
Extraction of 13 layers of convolutional features takes around 1 ms per frame.
Copying the data from GPU takes 10ms, which can be avoided by further
modifications to the software. Time spent on performing the alignment depends on
the number of iterations performed at different levels and usually is around
40--60ms.

\section{Results}

We present experiments which compare the performance of image alignment for our
CNN pyramid with raw RGB and dense SIFT pyramids. For all tests we use our
3D camera rotation tracker described in Section~\ref{sec:sphereslam}. We
consider that robustness is the most important factor in camera tracking, and
therefore use the size of basin of convergence as our performance measure in the main results in Section~\ref{sec:robust_tracking}. 

We also investigate the possibility of improving the results and reducing the computational overhead through selecting the most valuable feature maps in Section~\ref{sec:ablation}.

\subsection{Robust, long-term tracking} 
\label{sec:robust_tracking}
In order to test the robustness of the proposed representation in long-term
tracking scenarios, three time-lapse sequences from a static camera have been captured showcasing
real-world changing lighting conditions. The videos cover 8--10 hours of
outdoor scenes, and have been sub-sampled into 2 minute clips. 

In our tests we focus on two of the captured sequences -- {\em window} and {\em
home}. The first features a highly specular scene which undergoes major and
irregular lighting changes throughout the day with no major outliers. The {\em
home} sequence shows a relatively busy street with frequent outliers. This sequence contains less specular surfaces, the observed illumination
changes have more a global character.

We have selected three snapshots from each of these sequences containing
different lighting conditions, and evaluated the area of the convergence basin
while trying to align each possible pair of frames. To serve as a baseline
comparison, we also present the results of alignment using RGB (RGB) and dense
SIFT features (SIFT). 

In order to measure the size of the convergence basin, we segment the parameter
space into a regular grid and initialize the Lucas-Kanade algorithm at each
point. Next, we perform the optimization for 1000 iterations and inspect the
final result. We find that when convergence is successful, the final
accuracy is generally good, and therefore define that if the tracking
result is within 0.07 radians of ground truth, we mark
the tested point as belonging to the convergence basin. The marked points are
next used to calculate the total area of the convergence basin.

The results are presented in Figures~\ref{fig:square_window}
and~\ref{fig:square_home}. Five pyramids levels of SIFT and RGB have been used
in the tests to compare with the proposed CNN pyramid. The missing values were
duplicated in the plots so that it is possible to easily compare the
convergence basin areas of the corresponding image resolutions (e.g. RGB/SIFT
level 5 corresponds to CONV levels 11--13). Note that the LK alignment results
are not symmetric with regard to which image is used as the template and which
as the reference, as only the gradient of the template image is used, which
might have impact on performance under varying lighting conditions and blur.

\begin{figure*}
  \centering
  \newcommand{\imsize}{2.8cm}
  \includegraphics[width=\imsize,height=\imsize]{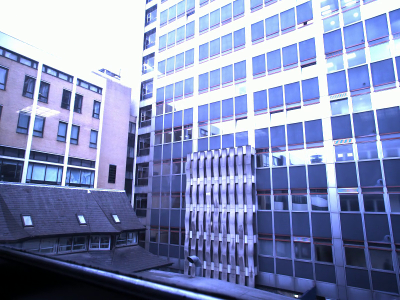}
  \includegraphics[width=\imsize,height=\imsize]{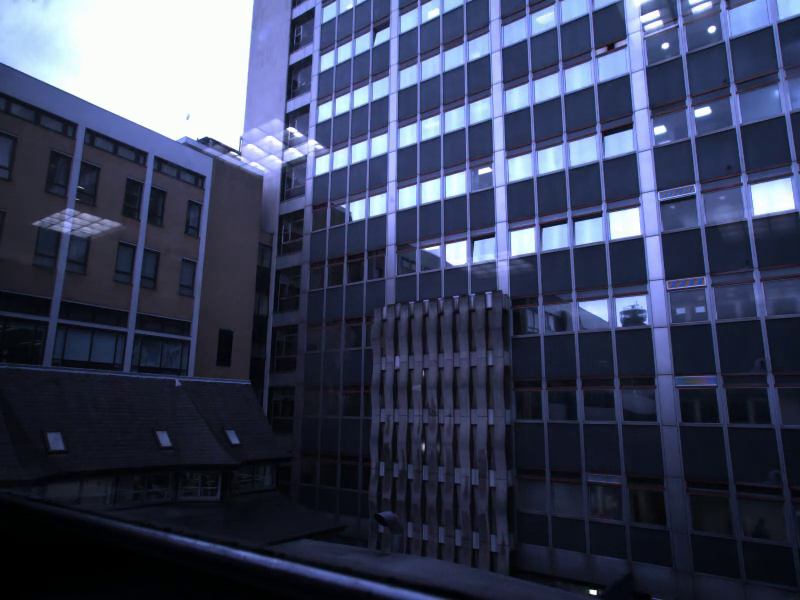}
  \includegraphics[width=\imsize,height=\imsize,clip,trim=7cm 4cm 8cm 4cm]{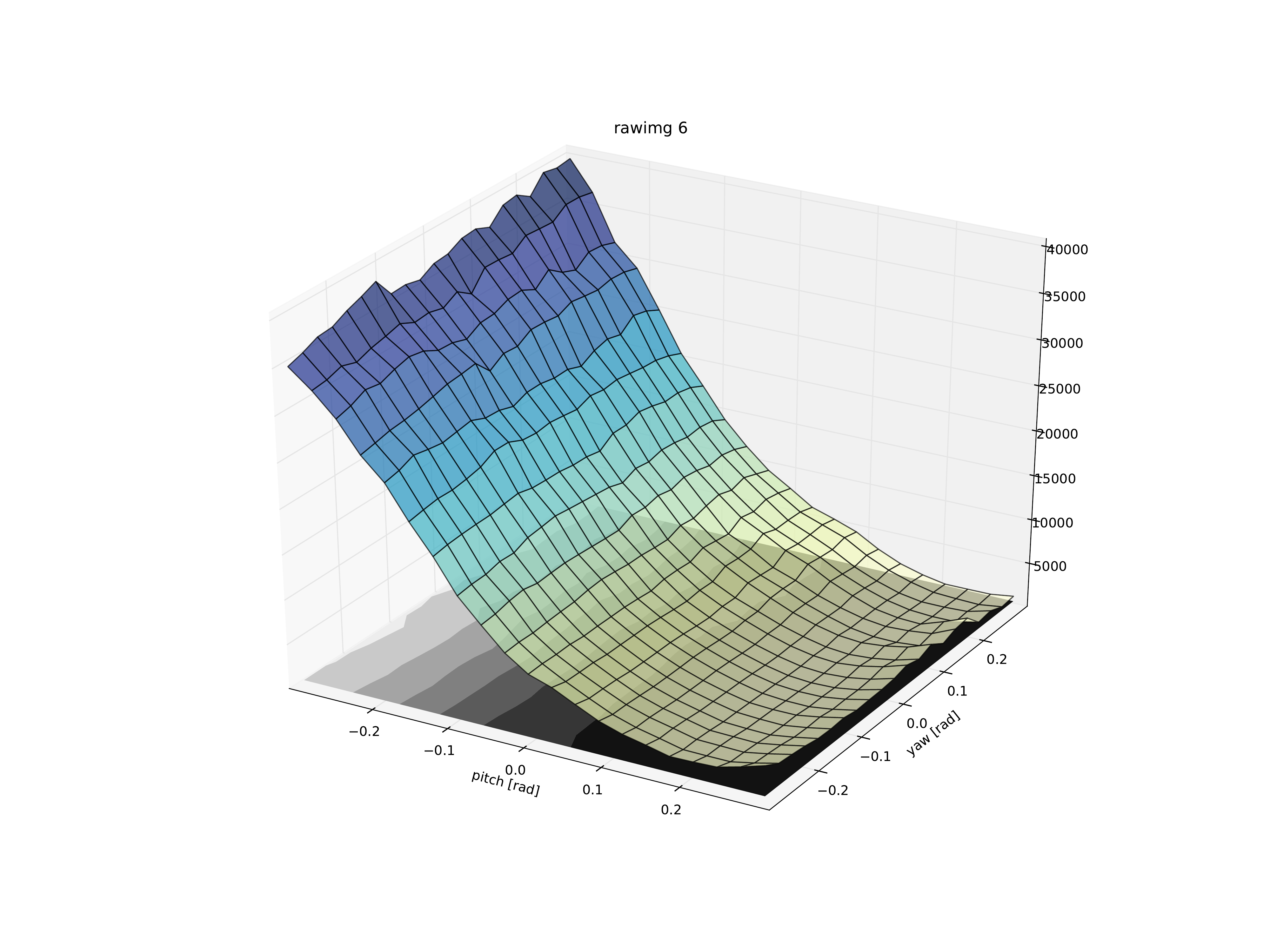}
  \includegraphics[width=\imsize,height=\imsize,clip,trim=5cm 4cm 8cm 4cm]{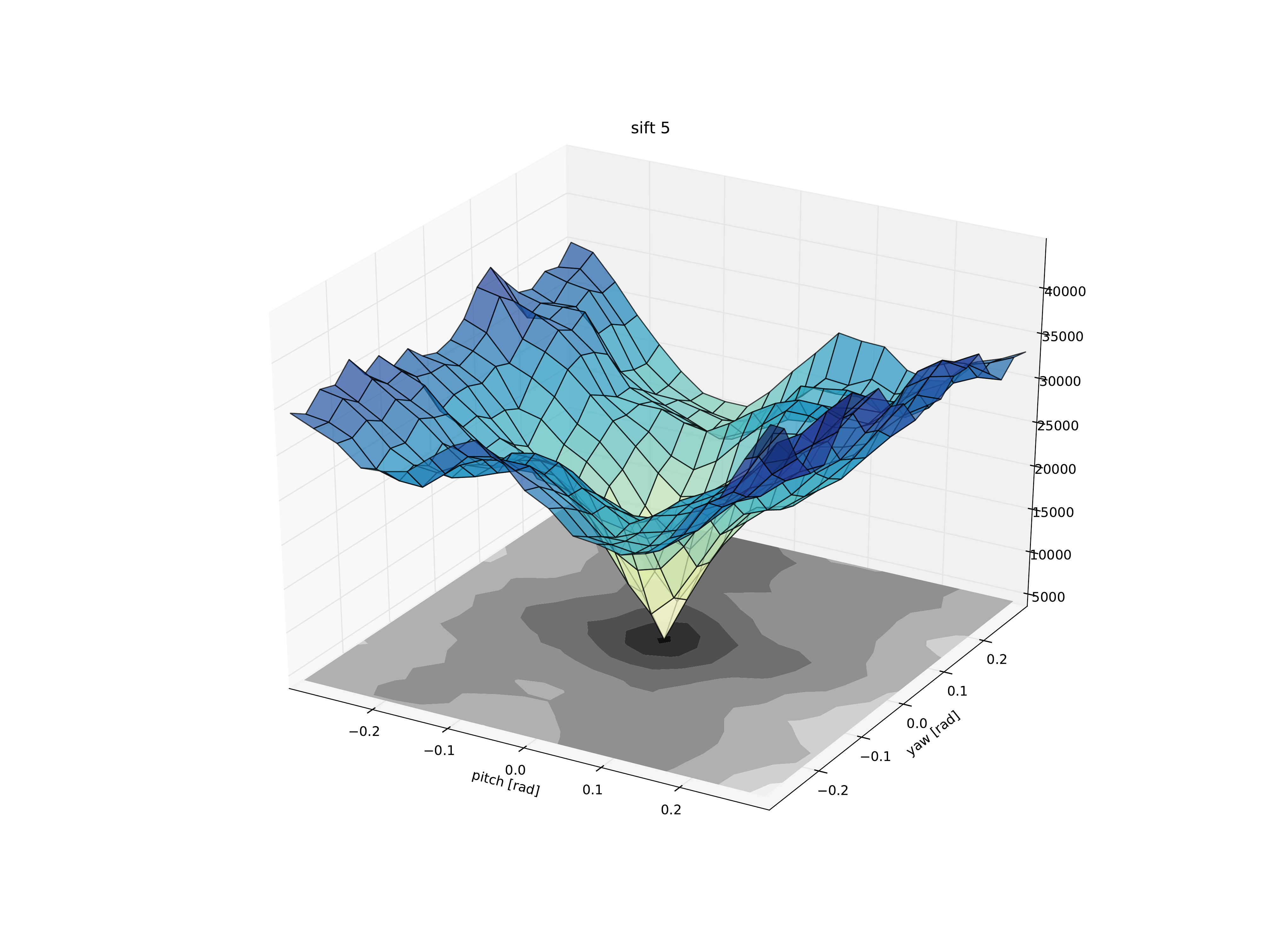}
  \includegraphics[width=\imsize,height=\imsize,clip,trim=6cm 4cm 8cm 4cm]{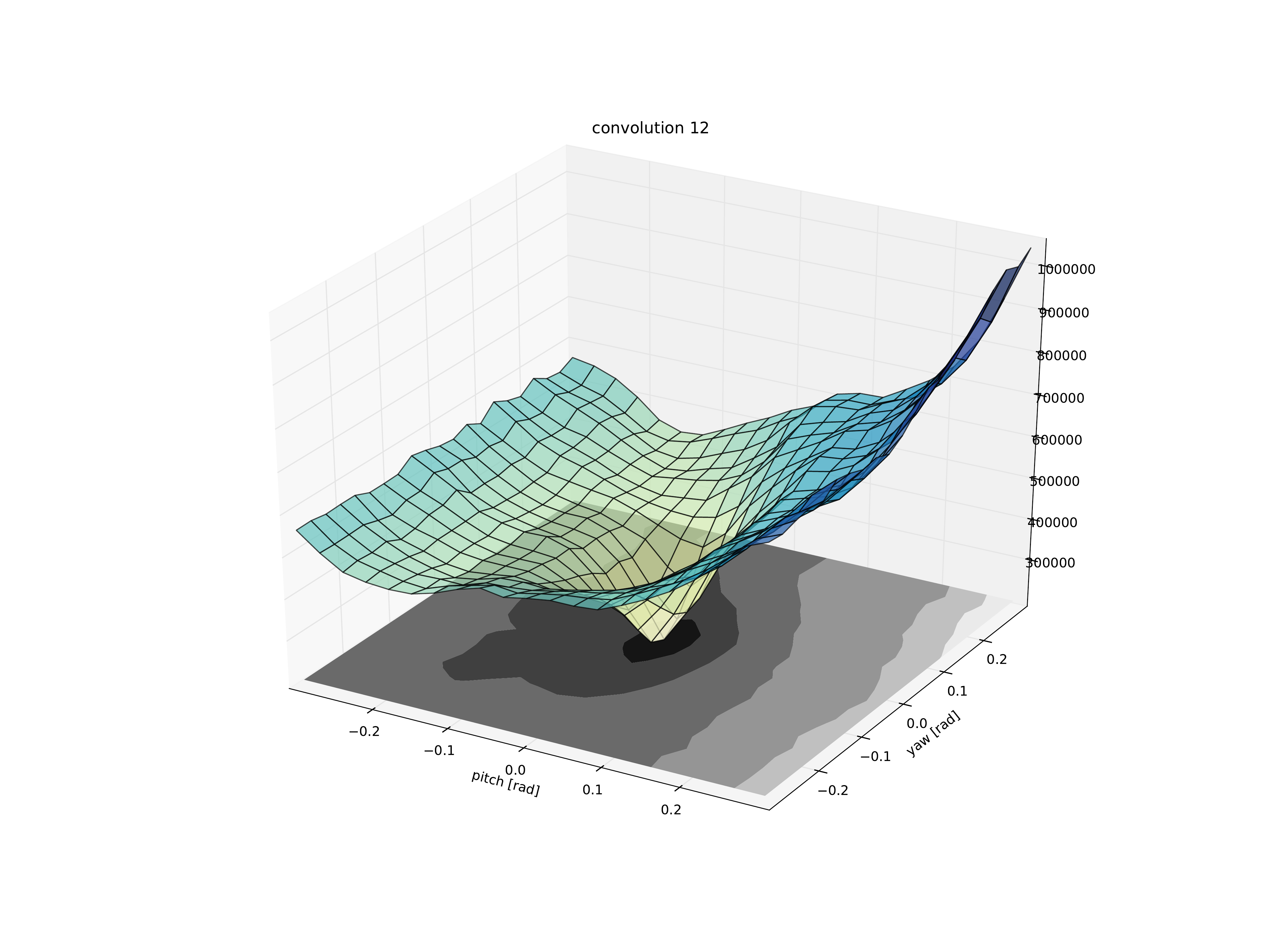}
  \includegraphics[width=\imsize,height=\imsize]{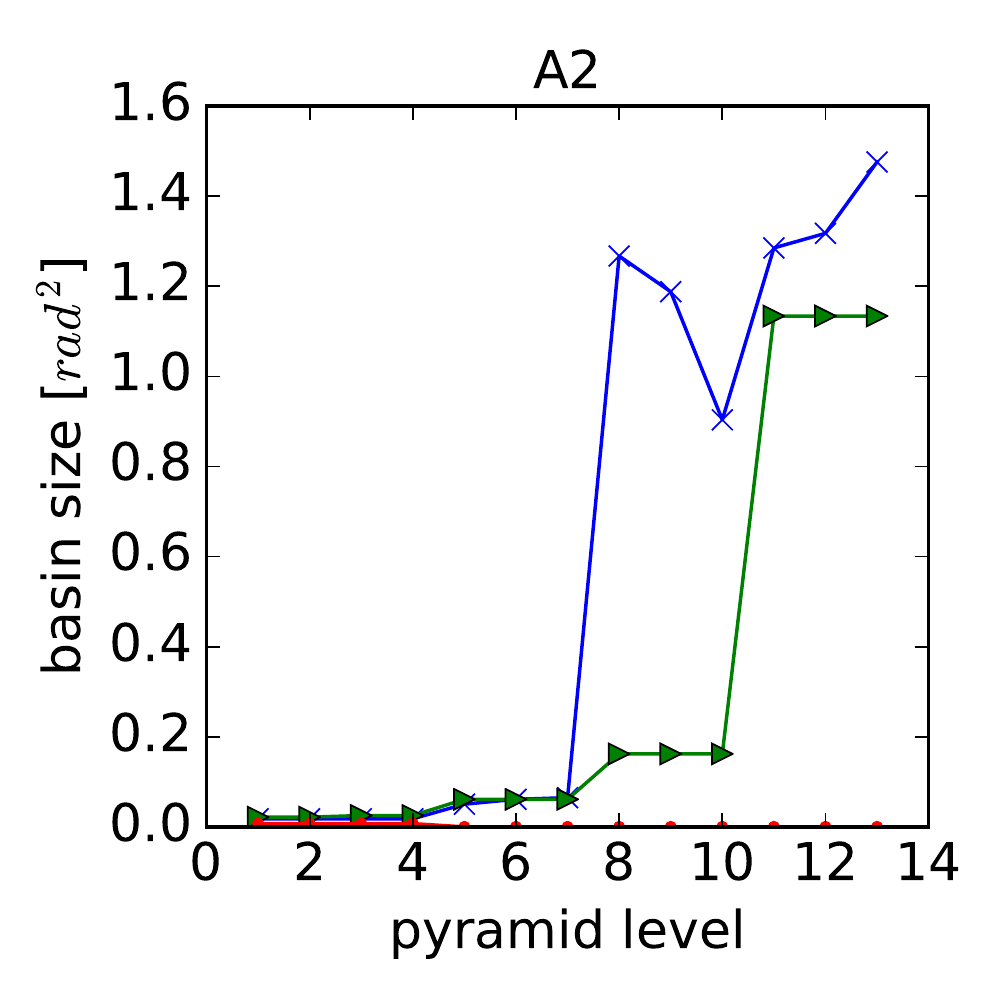}
  \caption{Cost landscape plots and convergence basin areas of different
  methods for aligning an image with a blurred version of itself. From left:
  template image, reference image, RGB cost landscape, SIFT cost landscape,
  CONV cost landscape, comparison of convergence basin areas (RGB: red, SIFT: green, CONV: blue).}
  \label{fig:rgb_fail}
\vspace{2mm}\hrule 
\end{figure*}

For frames from the {\em window} dataset (Figure~\ref{fig:square_window}), all of
the methods have a sensible basin of convergence on the diagonal, where the
images used for alignment are identical. For more challenging, off-diagonal
pairs, such as A2, A3 or B3, the RGB cost function fails to provide a minimum
in the correct spot, while SIFT and CONV still have wide, comparable
convergence basins.  One of the failure cases is showcased in more detail
in Figure~\ref{fig:rgb_fail}. The proposed CONV method seems to excel at higher
levels, which are believed to have more semantic meaning. 

In the second, {\em home} sequence (Figure~\ref{fig:square_home}) RGB performs
better, possibly due to the global character of the illumination changes. It
still fails when the lighting changes significantly (pair C2,B3).  Similar to
the previous sequence, the proposed method performs at least as well as SIFT
and better than RGB at the highest levels of pyramid. 

\begin{figure*}
  \centering
  \newcommand{\imsize}{2.8cm}
  \includegraphics[width=\imsize,height=\imsize]{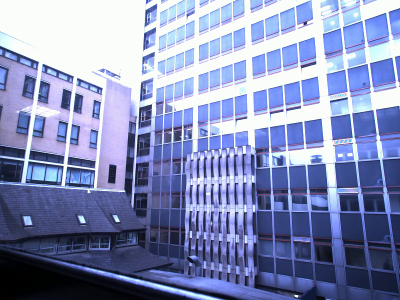}
  \includegraphics[width=\imsize,height=\imsize]{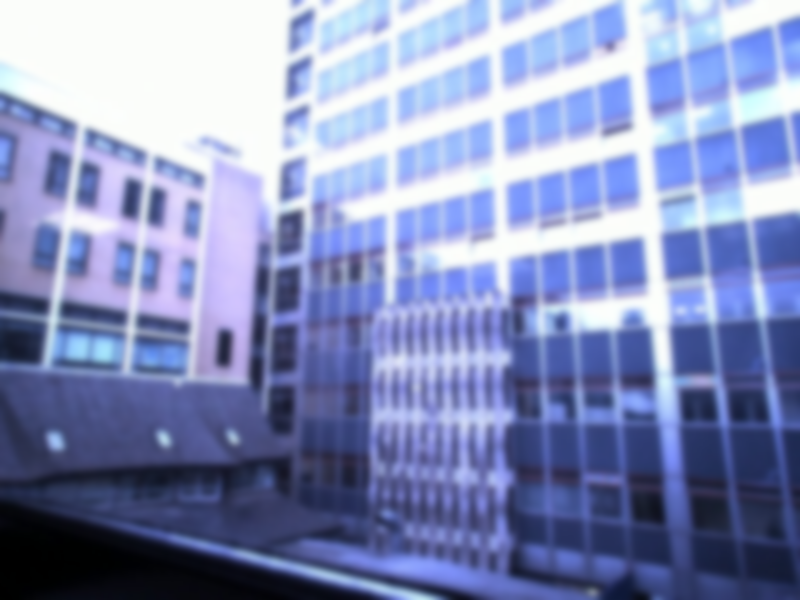}
  \includegraphics[width=\imsize,height=\imsize,clip,trim=6cm 4cm 8cm 4cm]{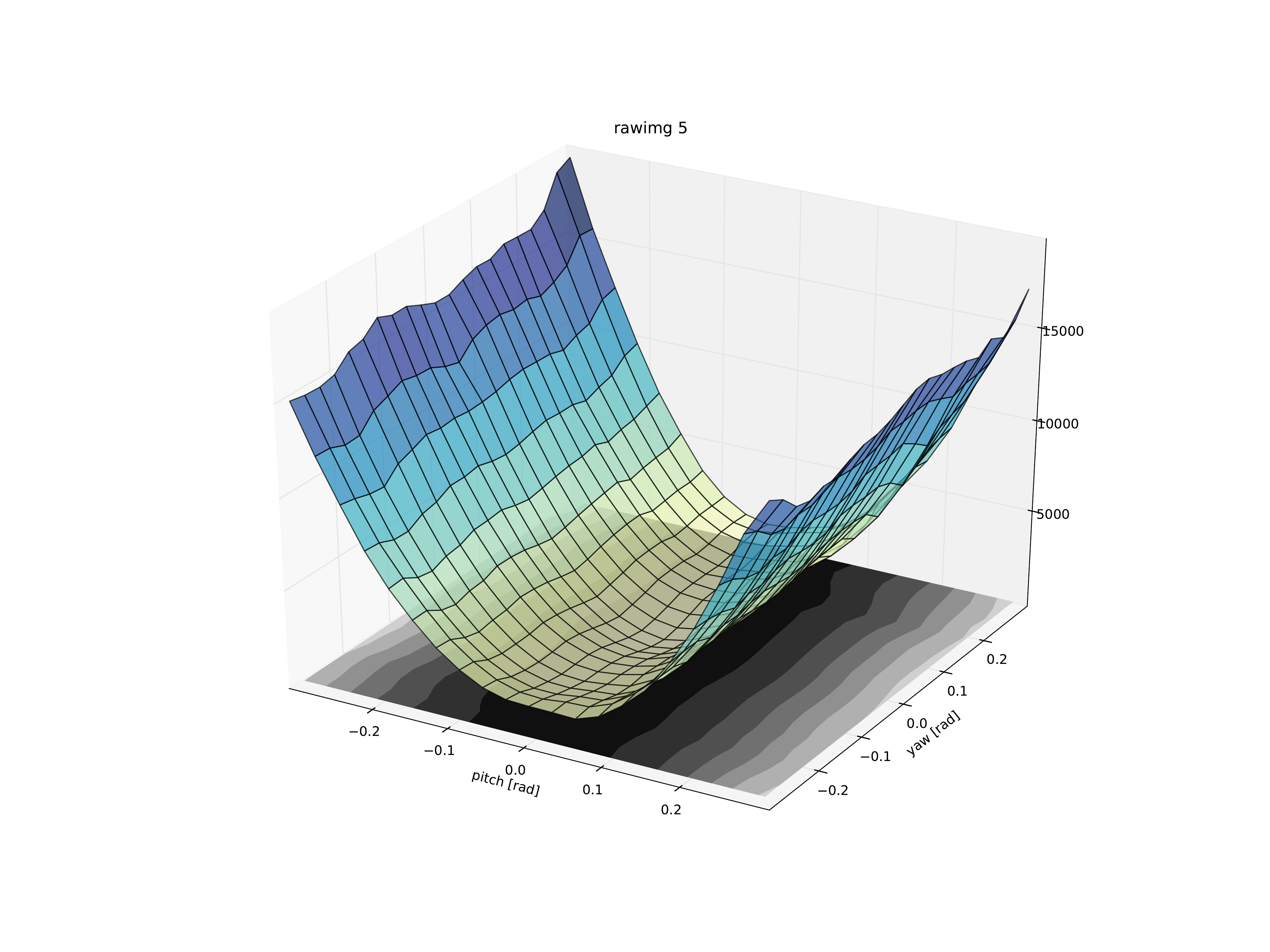}
  \includegraphics[width=\imsize,height=\imsize,clip,trim=6cm 4cm 8cm 4cm]{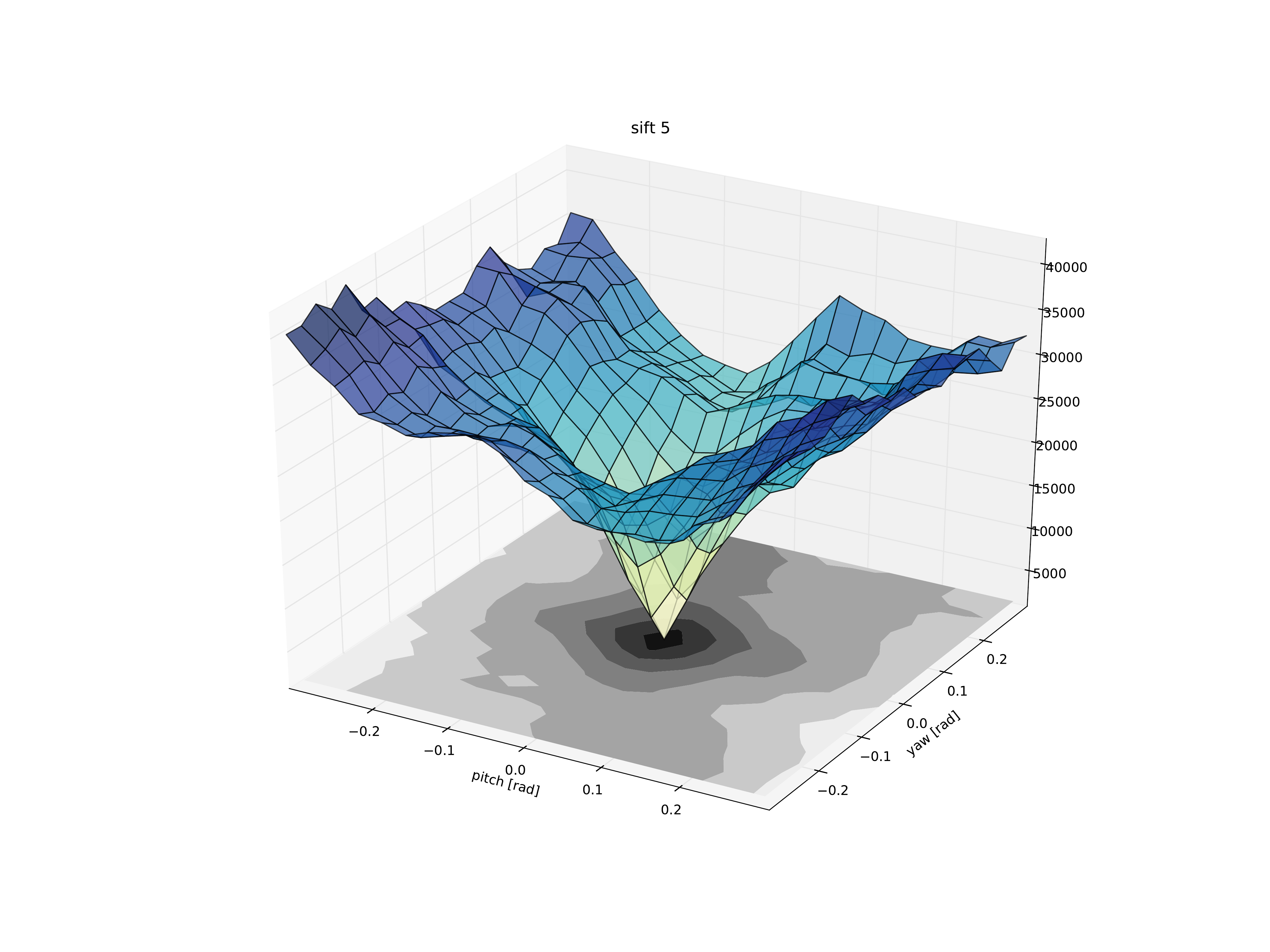}
  \includegraphics[width=\imsize,height=\imsize,clip,trim=6cm 4cm 8cm 4cm]{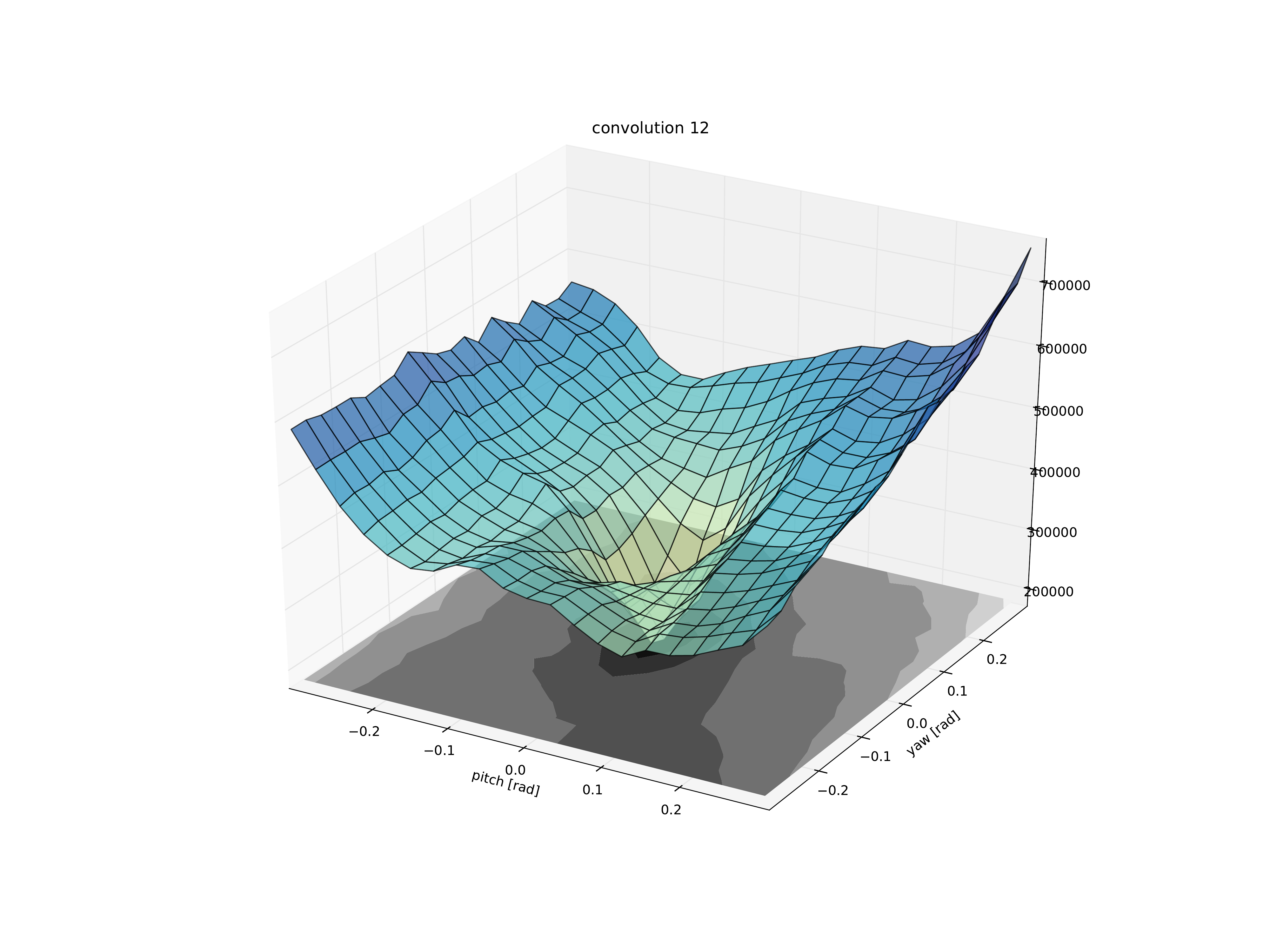}
  \includegraphics[width=\imsize,height=\imsize]{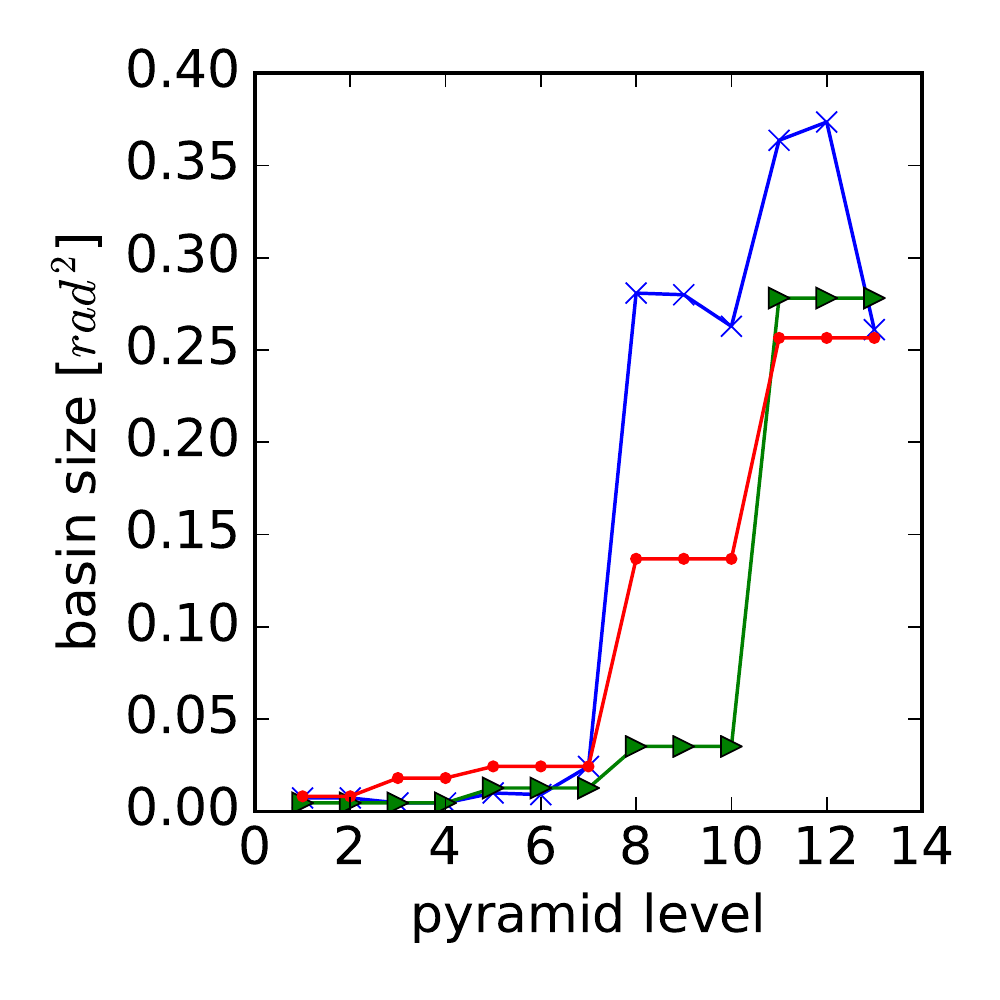}
  \caption{Cost landscape plots and calculated convergence basin areas of
  different methods for aligning an image with a blurred version of itself.
  From left: template image, reference image, RGB cost landscape, SIFT cost
  landscape, CONV cost landscape, comparison of convergence basin areas (RGB:
  red, SIFT: green, CONV: blue).}
  \label{fig:blur_case}
\vspace{2mm}\hrule 
\end{figure*}

To evaluate how the proposed solution handles image blurring, we have tested it
with an alignment problem of an image with a heavily blurred version of itself.
Figure~\ref{fig:blur_case} presents the reference and template images used in this
test, selected cost landscape plots of RGB, CONV and SIFT and a comparison of
basin sizes. We can see that all methods are robust to blur, with our proposed
method providing the best results at the highest pyramid levels.
It provides a steady, wide basin of convergence at
the top
pyramid levels regardless of the lighting conditions and blur.

\newcommand{\tabimgwidth}{3.9cm}

\newcommand{\tabimg}[2]{
  \includegraphics[width=\tabimgwidth,height=\tabimgwidth]{./data/squarefig/#1/#2}
}

\newcommand{\squarefig}[1]{
  \renewcommand{\unitlength}{1cm}
  \begin{tabular}{ c c  c  c  c  c||}
    & \tabimg{#1}{1.png} & \tabimg{#1}{2.png}& \tabimg{#1}{3.png} \\
    \tabimg{#1}{1.png} & \tabimg{#1}{1_1.pdf} &  \tabimg{#1}{1_2.pdf} & 
    \tabimg{#1}{1_3.pdf} \\
    \tabimg{#1}{2.png} & \tabimg{#1}{2_1.pdf} &  \tabimg{#1}{2_2.pdf} &
    \tabimg{#1}{2_3.pdf} \\
    \tabimg{#1}{3.png} & \tabimg{#1}{3_1.pdf} &  \tabimg{#1}{3_2.pdf} & 
    \tabimg{#1}{3_3.pdf} \\
  \end{tabular}
}

\begin{figure*}[t]
  \centering
  \renewcommand{\arraystretch}{0.5}
  \setlength\tabcolsep{1pt}
  \squarefig{window}
  \caption{Comparison of sizes of convergence basins for aligning pairs of
  images with different lighting conditions (sampled from the {\em window}
  sequence). Each array cell presents convergence basins areas of RGB (red),
  SIFT (green) and CONV (blue) at different pyramid levels. The left
  column and top row images
  are used in LK as template and reference, respectively.}
  \label{fig:square_window}
\vspace{2mm}\hrule 
\end{figure*}

\begin{figure*}
  \centering
  \renewcommand{\arraystretch}{0.5}
  \setlength\tabcolsep{1pt}
  \squarefig{home}
  \caption{Comparison of sizes of convergence basins for aligning pairs of
  images with different lighting conditions (sampled from the {\em home}
  sequence). Each array cell presents convergence basins areas of RGB (red),
  SIFT (green) and CONV (blue) at different pyramid levels. Left and top images
  are used in LK as template and reference, respectively.}
  \label{fig:square_home}
\vspace{2mm}\hrule 
\end{figure*}

\subsection{Reducing the Number of Features}
\label{sec:ablation}
The results presented so far are based on aligning all of the CNN feature maps
produced at each pyramid level jointly. One clear disadvantage of this approach
is computational cost: the amount of work which needs to be done at each
alignment iteration to calculate a difference is proportional to the number of
features. It is apparent from any inspection of the features generated by a CNN
that there is a lot of redundancy in feature maps, with many looking very
similar, and therefore it seemed likely that it is possible to achieve similar
or better results through selection of a percentage of features.
Here we perform experiments to compare a random selection with simple criteria
based on measures of texturedness  and stability. 

In standard Lucas-Kanade, the size of the convergence basin  depends
highly on image content. For example, a vertical stripe can only be
used to detect horizontal translation; the lack of vertical gradient
makes it impossible to determine the movement in this direction. In
order to correctly regress the camera pose, a strong gradient in both feature directions is required. As in Shi and Tomasi's classic `Good Features to Track' paper~\cite{Shi:Tomasi:CVPR1994}, we measure the texturedness of a feature based on its structure tensor:
\begin{equation*}
\label{eq:gradmat}
\matG_f = \sum_{\vecx \in \matI_f}
\begin{bmatrix}
g_x^2  & g_xg_y \\
g_xg_y & g_y^2
\end{bmatrix},
\end{equation*}
where $I_f$ is the activation map of the feature in response to a certain input image. We use the smallest eigenvalue of matrix $\matG_f$ as a comparable single score.

The other factor to consider is stability. Most valuable features provide stable activations that change only due to changes in camera pose. Features that react to objects that are likely to change their location, or lighting changes, are undesirable.
We measure the instability of a feature $f$ by calculating the average sum of squared differences (SSD) between activations obtained from images of the test sequence to the ones extracted from the first image and warped to the current camera frame:
\begin{equation*}
\label{eq:stability}
s_{f} = \frac{1}{N-1} \sum_{i=2}^{N}\sum_{\vecx} \norm{(\matI_f^1(\matW(\vecx;\vecomega)) - \matI_f^i(\vecx)}^2.
\end{equation*}
Averages of these two scores have been calculated across frames of three video sequences --- two timelapses and one hand tracking sequence. We have selected and evaluated several subsets of features of varying size that have the optimal texturedness and stability. To assess their robustness, we again use the size of convergence basin as a measure. Twelve challenging (outliers, varying lighting conditions) image pairs sampled from our recorded sequences have been used in the evaluation.

\begin{figure}
  \centering
  \includegraphics[width=\columnwidth,clip,trim=0cm 0cm 1cm 0.7cm]{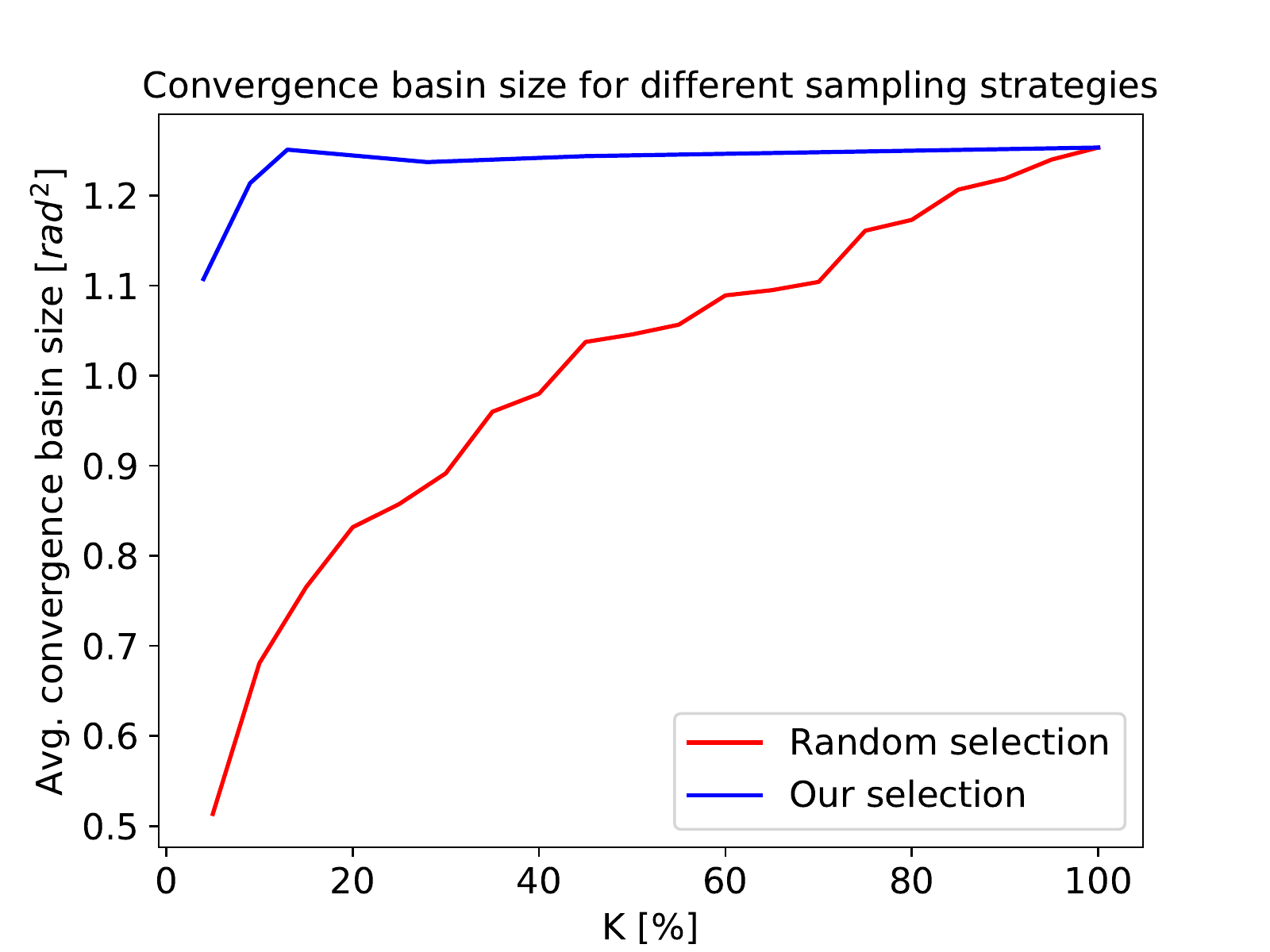}
  \caption{Average convergence basin size obtained by using our proposed selection process (blue) and random sampling (red), for different sampling percentage. Results were evaluated using twelve image pairs representing the most challenging conditions (outliers, lighting variations).}
  \label{fig:subsample_test}
\vspace{2mm}\hrule 
\end{figure}

Figure~\ref{fig:subsample_test} compares the average convergence basin size achieved with the features selected with our proposed approach with a baseline of using random selections. Random tests were performed in a range of 5--100\% subsampling, with a step of 5\%. Each sampled subset was tested 20 times against all twelve image pairs. We see very strongly that the features selected using our simple criteria give excellent tracking performance when they are much fewer than the whole set; while the performance of the randomly chosen features tails off much faster. This promises more sophisticated ways of learning ideal feature sets in the future.


\section{Conclusions}

We have shown that substituting the image pyramid in standard Lucas
Kanade dense alignment with a hierarchy of feature maps generated by a
standard VGG trained for classification straightforwardly gives much
improved tracking robustness with respect to large lighting changes.

While this is immediately a neat and convenient method for tracking,
we hope that it opens the door to further research on new
representations for dense mapping and tracking which lie in between
raw pixel values and object-level models. What levels of
representation are needed in dense SLAM in order actually to achieve
tasks such as detecting a change in an environment?
A clear next step would be to investigate the performance of
m-estimators on the CNN features to gate out outliers. Might we expect
to see that whole semantic regions which had moved or  changed could
be masked out from tracking? 


We imagine that a dense 3D model will be painted with smart learned
feature texture for tracking and updating, rather than the raw pixel
values of systems like DTAM~\cite{Newcombe:etal:ICCV2011}.

\section*{Acknowledgments}
Research presented in this paper has been supported by
Dyson Technology Ltd.
We thank Tristan Laidlow for help in capturing the datasets
used for evaluation in this paper.

{\small
\bibliographystyle{ieee}
\bibliography{robotvision}
}

\end{document}